\documentclass{acsa}
\usepackage[colorlinks,urlcolor=blue,linkcolor=blue,citecolor=blue]{hyperref}
\expandafter\def\expandafter\UrlBreaks\expandafter{\UrlBreaks\do\/\do\*\do\-\do\~\do\'\do\"\do\-}
\usepackage{color}
\usepackage{amsmath}
\newcommand{\furl}[1]{\footnote{\url{https://#1}}}

\usepackage{booktabs}
\usepackage{colortbl}
\usepackage{hyperref}
\usepackage{multirow}
\usepackage{tabularx}
\usepackage{bbm}
\usepackage{subfigure}

\let\<\textless
\let\>\textgreater

\begin{document}


\title{Exploring ChatGPT-based Augmentation Strategies for Contrastive Aspect-based Sentiment Analysis}

\author{~Lingling Xu}
\affil{~Hong Kong Metropolitan University, Hong Kong, China}

\author{~Haoran Xie, S. Joe Qin}
\affil{~Lingnan University, Hong Kong, China}

\author{~Fu Lee Wang}
\affil{Hong Kong Metropolitan University, Hong Kong, China}

\author{~Xiaohui Tao}
\affil{University of Southern Queensland, Queensland, Australia}


\begin{abstract}\looseness-1Aspect-based sentiment analysis (ABSA) involves identifying sentiment towards specific aspect terms in a sentence, and allows us to uncover people’s nuanced perspectives and attitudes on particular aspects of a product, service, or topic. However, the scarcity of labeled data poses a significant challenge to training high-quality models. To address this issue, we explore the potential of data augmentation using ChatGPT, a well-performing large language model (LLM), to enhance the sentiment classification performance towards aspect terms. Specifically, we explore three data augmentation strategies based on ChatGPT: context-focused, aspect-focused, and context-aspect data augmentation techniques. Context-focused data augmentation focuses on changing the word expression of context words in the sentence while keeping aspect terms unchanged. In contrast, aspect-focused data augmentation aims to change aspect terms but keep context words unchanged. Context-Aspect data augmentation integrates the above two data augmentations to generate augmented samples. Furthermore, we incorporate contrastive learning into the ABSA tasks to improve performance. Extensive experiments show that all three data augmentation techniques lead to performance improvements, with the context-aspect data augmentation strategy performing best and surpassing the performance of the baseline models.
\end{abstract}

\maketitle
\chapteri{A}spect-based sentiment analysis (ABSA) is a fine-grained sentiment classification task that focuses on discerning the sentiment polarity of aspect terms in a given sentence. Take the following sentence as an example: “This restaurant’s decor is eye-catching, but the food is unsatisfactory.” The aspect term “decor” exhibits a positive sentiment polarity, whereas the aspect term “food” demonstrates a negative sentiment polarity. Recently, the success of contrastive learning has led to the emergence of various data augmentation methods \cite{wang-etal-2022-contrastive,liang2021enhancing,xu2023improving}, which have been applied to contrastive aspect-based sentiment classification tasks, further improving the performance of ABSA tasks. Moreover, the development of LLMs, particularly the introduction of ChatGPT, has revolutionized the field of natural language processing (NLP). The advent of GPT-3.5 and GPT-4 has further expanded advanced applications across different domains, including ABSA tasks. However, directly using LLMs for fine-tuning with limited datasets poses challenges such as overfitting and expensive GPU costs. Therefore, a strategy to leverage the strong generative capabilities of LLMs is to use them for data augmentation. These augmented samples can then be employed in smaller models like BERT to undergo fine-tuning, thereby reducing computational demands and enhancing flexibility during fine-tuning.

Inspired by the work \cite{cheng-etal-2023-improving,ye2024llm}, which leverages ChatGPT for data augmentation in contrastive sentence representation learning and few-shot named entity recognition, we further investigate the role of ChatGPT for data augmentation in ABSA tasks. We propose three data augmentation strategies: context-focused, aspect-focused, and context-aspect data augmentation methods. In context-focused data augmentation, we aim to replace context words with other words while ensuring that the aspect terms and sentiment polarity remain unchanged. On the other hand, in aspect-focused data augmentation, we aim to replace aspect terms with other suitable aspect terms while preserving sentiment polarity. Finally, the context-aspect data augmentation combines both strategies mentioned above. Through comprehensive experiments, we observed that these three data augmentation techniques yielded improvements compared to the vanilla BERT, with context-aspect data augmentation demonstrating the best performance on both datasets. Furthermore, we found that employing data verification to ensure that generated aspect terms are completely different from the original ones does not consistently result in improved performance gains. Overall, the main contributions of this paper are as follows: 
\begin{itemize}
    \item We introduce how to prompt ChatGPT to perform context-focused and aspect-focused data augmentation for ABSA tasks.
    \item We explore the use of ChatGPT-based data augmentation and data verification strategies to boost the performance of ABSA tasks.
    \item We conduct exhaustive experiments to illustrate the effectiveness of ChatGPT for ABSA tasks in data augmentation.
\end{itemize}

\section{Related Work}
ABSA has gained a considerable amount of attention recently. A study from \cite{ma2018sentic} employed attention mechanisms, Long Short-Term Memory (LSTM), and external commonsense knowledge to enhance ABSA performance. One study found in \cite{ma2018targeted} introduced a hierarchical attention mechanism that combined context-focused and aspect-focused attention models. Another approach, Sentic GCN \cite{liang2022aspect}, utilized SenticNet\footnote{\url{https://sentic.net/}} to enrich the dependency graphs of sentences using graph convolution networks. By enhancing the dependencies between contextual words and aspect terms, this method aimed to capture affective information about aspect terms. The latest version, SenticNet 8 \cite{cambria2024senticnet}, further improves this by integrating Emotion AI and Commonsense AI to enhance interpretability, trustworthiness, and explainability in affective computing, which is crucial for tasks like ABSA. Furthermore, contrastive learning \cite{chen2020simple}, a self-supervised representation learning method, has received a great deal of attention and has been applied to various NLP tasks, including ABSA. This approach learns representations by bringing positive pairs (semantically similar sentences) closer together and pushing negative pairs (semantically different sentences) apart. For instance, some work \cite{wang-etal-2022-contrastive,xu2023improving} proposed novel data augmentation strategies to increase the number of training datasets to improve ABSA through contrastive learning. More recently, the remarkable performance and powerful generative and rewriting capabilities of LLMs like ChatGPT have inspired researchers to explore their potential for data augmentation. In this regard, \cite{cheng-etal-2023-improving} leverages ChatGPT and word masking to generate new sentences and improve contrastive sentence representation learning. While \cite{ye2024llm} utilizes prompt and ChatGPT for data augmentation to improve the performance of few-shot named entity recognition.

\section{Methodology}
In this section, we begin by introducing the ABSA task. Next, we present three novel data augmentation methods based on ChatGPT for the ABSA datasets. Lastly, we outline the overall training objective of our proposed model framework.

\begin{figure*}
\centering
\includegraphics[width=\linewidth]{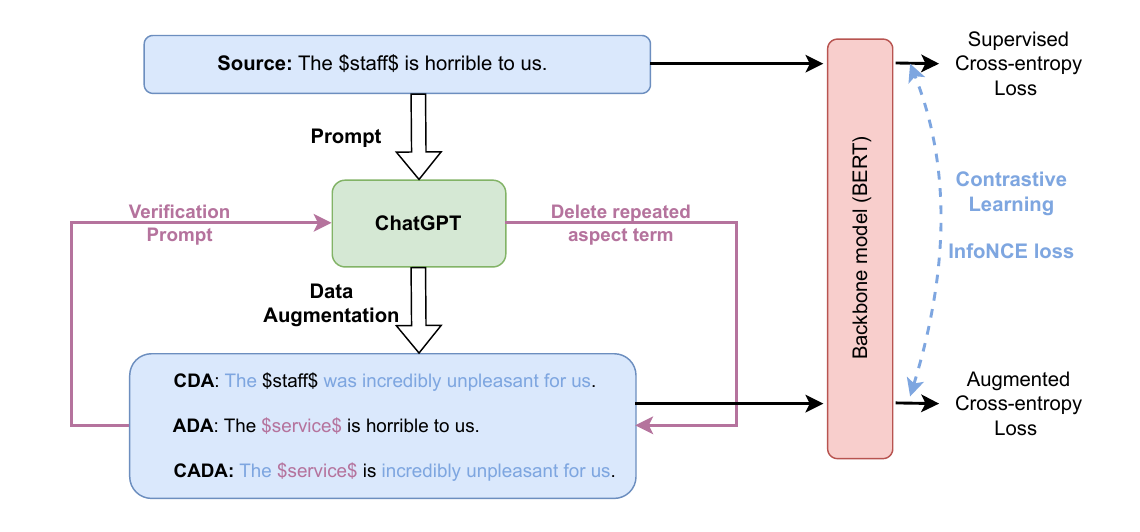}
\caption{The overall framework of our proposed method.}
\label{fig1}
\end{figure*}

\subsection{Task Formulation}
ABSA datasets are comprised of a group of triplet samples $\mathbf{D} = \{s_{i}, a_{i}, l_{i}\}_{i=1}^{N}$, in which $s_{i}$ denotes review sentences, $a_{i}$ means aspect term, and $l_{i} \in \{-1, 0, 1\}$ denotes the sentiment polarity towards aspect term (\emph{-1 denotes negative polarity, 0 denotes neutral polarity, 1 denotes positive polarity}). Following the work \cite{liang2021enhancing,li-etal-2021-learning-implicit}, we also choose the pretrained language model BERT \cite{devlin-etal-2019-bert} as the backbone model and construct sentence pair classification task to obtain the sentiment representations towards the aspect term in the given sentence. More specifically, the inputs are first transformed into “[CLS]s$_{i}$[SEP]a$_{i}$[SEP]” with [CLS] and [SEP] tokens. These transformed inputs are then fed into BERT to obtain the aspect-based sentiment representation:
\begin{equation}
    h_{i} = \mathcal{F}([CLS]s_{i}[SEP]a_{i}[SEP]),
\end{equation}
in which $\mathcal{F}(\cdot)$ denotes the BERT model. 

\subsection{Data Augmentation}
We present three ChatGPT-based data augmentation methods for ABSA datasets with the assistance of GPT-3.5-turbo, which was developed by OpenAI\footnote{\url{https://platform.openai.com/docs/models/gpt-3-5-turbo}}. Moreover, some examples are presented in Table~\ref{tab1} to illustrate the distinctions among three data augmentation strategies.

\begin{table*}
\centering
\caption{Comparing the generated samples with CDA, ADA, and CADA methods, the colored spans are changed part.}
\begin{tabular*}{35.8pc}{@{}cc@{}}
\toprule
\textbf{Method}  & \textbf{Examples [dataset, aspect term, sentiment]}\\ 
\midrule
\textbf{Source} & The speed is incredible and I am more than satisfied. \textcolor{blue}{[Laptop, speed, positive]} \\
\textbf{CDA} &   \textcolor{cyan}{The} speed \textcolor{cyan}{is extraordinary and I am more than content}. \\
\textbf{ADA} &  The \textcolor{red}{performance} is incredible and i am more than satisfied.   \\
\textbf{CADA} &  \textcolor{cyan}{The} \textcolor{red}{performance} \textcolor{cyan}{is extraordinary and I am more than content}.   \\
\midrule
\textbf{Source} & The palak paneer was standard, and I was not a fan of the malai kofta. \textcolor{blue}{[Restaurant, palak paneer, neutral]} \\
\textbf{CDA}  &   \textcolor{cyan}{The} palak paneer \textcolor{cyan}{was mediocre, and I did not enjoy the creamy vegetable balls}. \\
\textbf{ADA}  &  The \textcolor{red}{curry} was standard, and I was not a fan of the malai kofta.  \\
\textbf{CADA} & \textcolor{cyan}{The} \textcolor{red}{curry} \textcolor{cyan}{was mediocre, and I did not enjoy the creamy vegetable balls}.  \\

\bottomrule
\end{tabular*}
\label{tab1} 
\end{table*}

\subsubsection{Context-focused Data Augmentation (CDA)} is a data augmentation technique that aims to change the contextual words in a sentence while keeping the aspect term and sentiment polarity of the aspect term unchanged. The goal of CDA is to increase the semantic richness and diversity of the training dataset while keeping the semantics of the sentences unchanged. Specifically, we adopt paraphrasing as the data augmentation strategy and utilize the excellent rewriting abilities and comprehensive world knowledge of ChatGPT for CDA to obtain augmented samples. However, constructing a suitable prompt for CDA is challenging, as it involves preserving aspect terms and their sentiment polarity unchanged, making it difficult to design a suitable prompt. Following the rule of principled guidance, the prompt for CDA is presented in Table~\ref{tab2}. Notably, two examples are presented in the prompt to facilitate ChatGPT’s comprehension of the intended task and generate the desired sentence. However, the selection of two examples varies to account for the differences in domain within the training datasets.

\begin{table}[htb]
\centering
\caption{ChatGPT prompt for ADA. Place the source sentence to the “\{sentence\}” and place the aspect term in the source sentence to the “\{aspect term\}”.}
\renewcommand{\arraystretch}{1.2}
\small
\begin{tabular}{p{6.8cm}}
\hline
Given the sentence: “\{sentence\}”, and given the aspect term “\$\{aspect term\}\$” in above sentence.\\
\\
Please generate one new sentence using paraphrasing. The new sentence should not paraphrase the aspect term “\$\{aspect term\}\$” and should keep the aspect term “\$\{aspect term\}\$”, semantics of the sentence, and sentiment polarity towards the aspect term “\$\{aspect term\}\$” unchanged. \\
\\
Here are a few examples: \\
Source sentence: \<source sentence\> $\rightarrow$ New sentence: \<CDA augmented sentence\> \\
\\
Please only output the New sentence.\\
\hline
\end{tabular}
\label{tab2} 
\end{table}


\subsubsection{Aspect-focused Data Augmentation (ADA)} is a data augmentation method that focuses on replacing the original aspect term with a different semantically and logically suitable aspect term. At the same time, it preserves the context words in the sentence, and the sentiment polarity towards the new aspect term is unchanged. The purpose of ADA is to increase the diversity of aspect terms and improve the robustness of the model in detecting the opinion words associated with different aspect terms. Unlike existing approaches that generate new sentences based on given aspect terms in \cite{wang-etal-2022-contrastive}, our approach employs ChatGPT to randomly generate new aspect terms, thereby increasing the generality and diversity of aspect terms, especially for unseen aspect terms. To perform CDA for ABSA datasets, the ChatGPT prompt for CDA is presented in Table~\ref{tab3}. Importantly, we have implemented a verification step to ensure that the newly generated aspect term differs from the original. In cases where the aspect term remains the same as the original, the source sentence is repeatedly fed back into ChatGPT with the same prompt until a distinct aspect term is generated.

\begin{table}[htb]
\centering
\caption{ChatGPT prompt for ADA. Place the source sentence to the “\{sentence\}” and place the aspect term in the source sentence to the “\{aspect term\}”.}
\renewcommand{\arraystretch}{1.2}
\small
\begin{tabular}{p{6.8cm}}
\hline
Given the sentence: “\{sentence\}”, and given the aspect term “\$\{aspect term\}\$” in above sentence. \\
\\
Please replace the given aspect term in the given sentence with a new semantically and logically suitable aspect term and also keep the sentiment polarity towards the new aspect term unchanged. \\
\\
Please only output the new aspect term. \\
\hline
\end{tabular}
\label{tab3} 
\end{table}

\subsubsection{Context-Aspect Data Augmentation (CADA)} is a data augmentation method that combines the aforementioned two data augmentation strategies. Specifically, CADA first utilizes the CDA on the source sentence to generate a context-focused augmented sentence, and then leverages ADA on the same source sentence to obtain a new aspect term that aligns with the semantics and logic of the source sentence. The augmented samples using CADA are achieved by concatenating the context-focused augmented sentence with the newly generated aspect term. By integrating CDA and ADA in this manner, CADA not only helps to diversify the sentence structure and wording, but also the aspect terms.

\subsection{Training Objective}
In this section, we introduce the overall training objective of our proposed method. The overall framework is shown in Figure 1. The overall training objective is to perform aspect-based supervised sentiment classification on both source and augmented samples, and to conduct contrastive learning between source samples and augmented samples. Specifically, we perform supervised sentiment classification training (SSCT) on source and augmented sentences with the following objectives:

\begin{equation}
\begin{split}
 \mathcal{L}_{SSCT} & = \frac{1}{N} \sum_{i=1}^{N} (L_{CE}(h_{i}W_{s} + b_{s}, l_{i}) \\
                    & + \alpha L_{CE} (h_{i}^{+}W_{s} + b_{s}, l_{i})),  
\end{split}
\end{equation}
in which $h_{i}$ and $h_{i}^{+}$ are the aspect-based sentiment representations of source samples and augmented positive samples, and $l_{i}$ is the ground-truth sentiment polarity label of source and augmented samples. $\alpha$ is the hyper-parameter to adjust the performance of the SSCT task.

To further enhance the performance robustness, we incorporate contrastive learning into our approach. In this method, the samples generated through the ChatGPT-based data augmentation strategy are considered positive pairs, while the rest of samples in the batch are regarded as negative pairs. The InfoNCE loss is used as our contrastive learning loss:
\begin{equation}
    \mathcal{L}_{CL} = - \frac{1}{N} \sum_{i=1}^{N} \log \frac{\exp(\cos(h_{i}, h_{i}^{+})/\tau)}{\sum_{j=1}^{N} \exp(\cos(h_{i}, h_{j}^{+})/\tau)}.
\end{equation}
Here, $N$ is the number of samples in a batch, with $\tau$ being the temperature parameter to adjust the similarity between source samples and augmented samples.

Thus, the overall training objective of our proposed framework can be formulated as:
\begin{equation}
    \mathcal{L} = \mathcal{L}_{SSCT} + \beta \mathcal{L}_{CL}
\end{equation}
in which $\beta$ is a controllable parameter.

\section{Experiments}
In this section, we begin by introducing training datasets and baseline models. Subsequently, we provide a comprehensive overview of our experimental settings and details. Additionally, we present and analyze the experimental results, and conduct sensitivity analysis to examine the impact of hyper-parameters on ABSA performance.

\subsection{Dataset} 
We conduct experiments on the two public ABSA datasets, Restaurant and Laptop, which are sourced from SemEval 2014 \cite{pontiki2016semeval}, comprising reviews in the domains of restaurants and laptops, respectively. The sample in the two ABSA datasets consists of a sentence, an associated aspect term, and the sentiment label towards the aspect term. The statistics and details of the two datasets are shown in Table~\ref{tab4}.

\begin{table}
\centering
\caption{Statistics of Laptop and Restaurant datasets.}
\begin{tabular*}{17.6pc}{@{}cllllll@{}}
\toprule
\multirow{2}{*}{\textbf{Dataset}} & \multicolumn{2}{c}{\textbf{Positive}} & \multicolumn{2}{c}{\textbf{Neutral}} & \multicolumn{2}{c}{\textbf{Negative}}\\ \cline{2-7}
& Train & Test & Train & Test & Train & Test \\ 
\midrule
\multicolumn{1}{l}{Restaurant} & 2164 & 728  & 637 & 196 & 807 & 196 \\ 
\multicolumn{1}{l}{Laptop} & 994 & 341 & 464 & 169 & 870 & 128 \\
\bottomrule
\end{tabular*}
\label{tab4} 
\end{table}

\subsection{Baselines} To evaluate the effectiveness of our proposed data augmentation approach, we compared it against four baseline models: 1) \textbf{BERT-base} directly uses the vanilla BERT model as the backbone and fine-tune BERT on the source training samples for aspect-based sentiment classification; 2) \textbf{C$^{3}$DA} \cite{wang-etal-2022-contrastive} proposes a contrastive cross-channel data augmentation method to obtain augmented samples for supervised aspect-based sentiment classification and contrastive learning;  3) \textbf{BERT-Scon} \cite{liang2021enhancing} leverages the characteristic of aspect-invariant and sentiment-invariant for supervised contrastive learning to enhance ABSA; 4) \textbf{BERT+SR} \cite{xu2023improving} utilizes synonym replacement for data augmentation to obtain augmented samples for contrastive aspect-based sentiment classification.

\subsection{Implementation Details} 
We implemented our experiments using the pretrained language model BERT (bert-base-uncased) as the backbone model. The batch size was set to 32, and we employed Adam as the optimizer with a learning rate of 2e-5 and a dropout rate of 0.1. Additionally, we set the temperature parameter to 0.08 for contrastive learning. Regarding data augmentation, we utilized the GPT-3.5-turbo model as the ChatGPT, setting the temperature parameter to 0 for CDA and 1 for ADA, aiming to preserve the sentence semantics while diversifying the aspect terms. For the hyper-parameters $\alpha$ and $\beta$ in the overall loss function, we adapted different values for each data augmentation strategy for better ABSA performance. Specific details can be found in Table~\ref{tab5}. It was discovered that CADA achieves better performance when selecting a larger hyper-parameter. Furthermore, all models were trained for 50 epochs, employing an early-stop strategy that terminates training if there are no improvements within 10 epochs. All of our experiments were conducted with a single NVIDIA A800 (80G).

\begin{table}
\centering
\caption{The hyper-parameters setting for different data augmentation strategies. (veri) denotes to delete the repeated aspect terms}
\begin{tabular*}{9.5pc}{@{}lccc@{}}
\toprule
\textbf{Dataset} & \textbf{$\alpha$} & & \textbf{$\beta$} \\
\midrule
CDA & 0.2 & & 0.2 \\
ADA & 0.6 & & 0.5 \\
ADA (veri) & 0.1 & & 0.2 \\
CADA & 0.2 & & 0.4 \\
CADA (veri) &  0.4 & & 0.6 \\
\bottomrule
\end{tabular*}
\label{tab5} 
\end{table}

\subsection{Main Results}
\begin{table*}
\centering
\caption{Experimental results with different data augmentation strategies on the ABSA datasets.}
\begin{tabular*}{23.3pc}{@{}lcccccccccc@{}}
\hline
\multirow{2}{*}{\textbf{Model}} &  &  & \multicolumn{3}{c}{\textbf{Restaurant}} & & & \multicolumn{3}{c}{\textbf{Laptop}} \\ \cline{4-6}\cline{9-11} 
&  &   & Acc. & & F1  & & & Acc. & & F1 \\ \hline
\multicolumn{1}{l}{BERT-base} & &  &  85.98 & & 79.84 & & & 79.47 & & 74.65  \\
\multicolumn{1}{l}{BERT+SR \cite{xu2023improving}} & &  &  85.80 & & 79.70 & & & 80.56 & & 76.11  \\
\multicolumn{1}{l}{BERT-Scon \cite{liang2021enhancing}} & &  &  86.51 & & 80.55  & & & 80.23 & & 76.48  \\
\multicolumn{1}{l}{C$^{3}$DA \cite{wang-etal-2022-contrastive}} &  &  & 86.93 & & 81.23  & & & 80.61 & &77.11 \\
\hline
\multicolumn{1}{l}{BERT+CDA} &  &  &  86.52 & & 80.36  & & & 80.56 & & 77.51  \\
\multicolumn{1}{l}{BERT+ADA} &  &  &  86.61 & & 80.91  & & & 80.25 & & 75.80  \\
\multicolumn{1}{l}{BERT+ADA (veri)} &  &  &  86.70 & & 80.27 & & & 80.25 & & 76.65 \\
\multicolumn{1}{l}{BERT+CADA} &  &  &  \textbf{87.14} & & \textbf{81.26}  & & & \textbf{80.88} & & \textbf{77.71}   \\
\multicolumn{1}{l}{BERT+CADA (veri)} &  &  &  86.88 & & 81.00 & & & 80.56 & & 77.09 \\

\hline
\end{tabular*}
\label{tab6}
\end{table*}

As indicated in Table~\ref{tab6}, ChatGPT-based augmentation techniques, including CDA, ADA, and CADA, consistently and significantly enhance performance across ABSA tasks. Notably, the BERT+CADA model achieved the highest performance in the Laptop and Restaurant datasets, surpassing all baseline models. The Laptop dataset demonstrated a remarkable improvement of 1.41$\%$ in accuracy and 3.06$\%$ in Macro F1 compared to the vanilla BERT model. Similarly, the Restaurant dataset showcased a notable enhancement of 1.16$\%$ in accuracy and 1.42$\%$ in macro F1 compared to the vanilla BERT model. Moreover, the incorporation of aspect term verification, utilizing GPT-3.5-turbo to eliminate repeated aspect terms, contributed to further accuracy improvements in the Restaurant dataset and macro F1 improvements in the Laptop dataset. We discovered that BERT+CADA (veri) outperformed BERT+CADA on both datasets. This may be due to the following reasons: first, repeated aspect terms are more in line with the semantics of the source sentence, and forcing the generation of different aspect terms is not semantically and logically appropriate for the sentence; second, certain repeated aspect terms in the CADA enhancement ensure that the semantic changes are not too large, leading to better performance. Additionally, BERT+CADA (veri) demonstrated superior performance compared to BERT+CDA and BERT+ADA in the Restaurant dataset, and better performance than BERT+ADA and BERT+ADA (veri) in the Laptop dataset. Unlike verification techniques used in LLMs to tackle hallucination and ensure accurate responses \cite{dhuliawala2023chain}, our verification step concentrated on generating distinct aspect terms. Significantly, we discovered that generating unsuitable, yet distinct aspect terms may potentially impact the semantics of sentences and result in degraded performance.

\subsection{Sensitivity Analysis}

As mentioned above, two hyper-parameters, $\alpha$ and $\beta$, were adopted in our proposed contrastive aspect-based sentiment classification framework. To investigate the influence of both hyper-parameters, a sensitivity analysis was conducted. We explored a range of values for $\alpha$ and $\beta$ in $\{0.1, 0.2, 0.3, 0.4, 0.5, 0.6,0.7, 0.8, 0.9, 1.0\}$, and performed a grid search to identify suitable combinations that yielded decent performance on both the Laptop and Restaurant datasets. As depicted in Figure 2, we observed consistent trends in the Accuracy and F1 scores across different hyper-parameter values. The supervised aspect-based sentiment classification performance exhibited sensitivity to $\alpha$, while the introduction of augmented cross-entropy loss yielded only marginal improvements, even a decrease, on the Laptop dataset across three data augmentation methods. However, notable improvements were observed in the Restaurant dataset. Additionally, as shown in Figure~\ref{fig3}, we also investigated the impact of hyper-parameter $\beta$ in the total loss function. We discovered that the overall performance of three different data augmentation strategies is sensitive to the hyper-parameter $\beta$. To ensure that our proposed framework achieves excellent performance in both the Laptop and Restaurant datasets across different methods, we employed distinct combinations of hyper-parameters for different data augmentation techniques, which are detailed in Table~\ref{tab5}.

\begin{figure*}
\centering
\subfigure[\scriptsize{Res-CDA}]{
\includegraphics[width=0.92in]{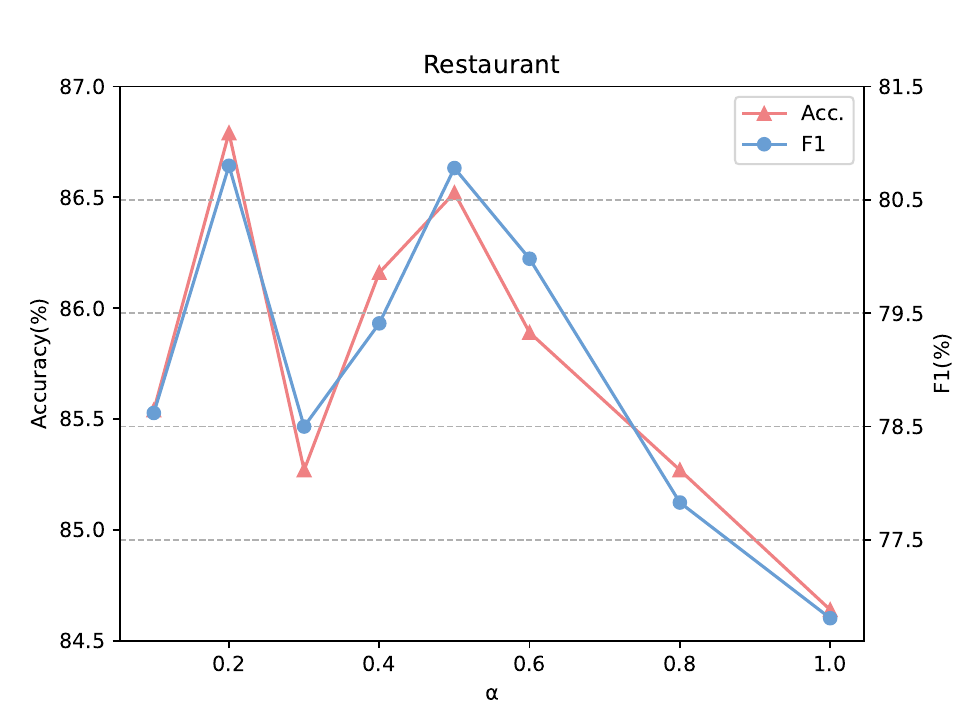}
}
\quad
\subfigure[\scriptsize{Res-ADA}]{
\includegraphics[width=0.92in]{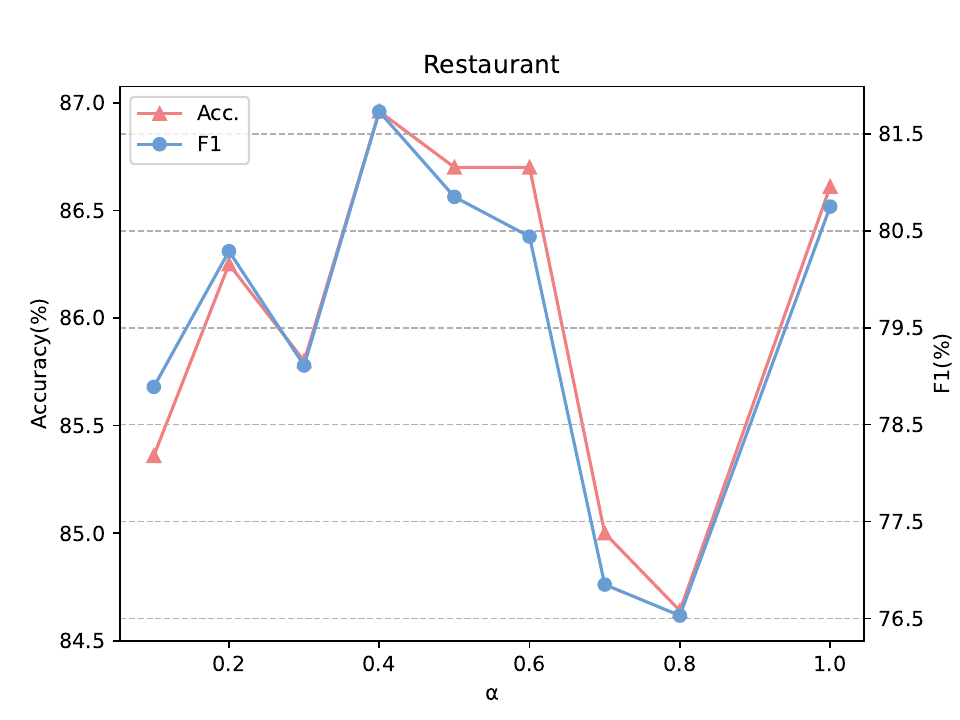}
}
\quad
\subfigure[\scriptsize{Res-ADA (veri)}]{
\includegraphics[width=0.92in]{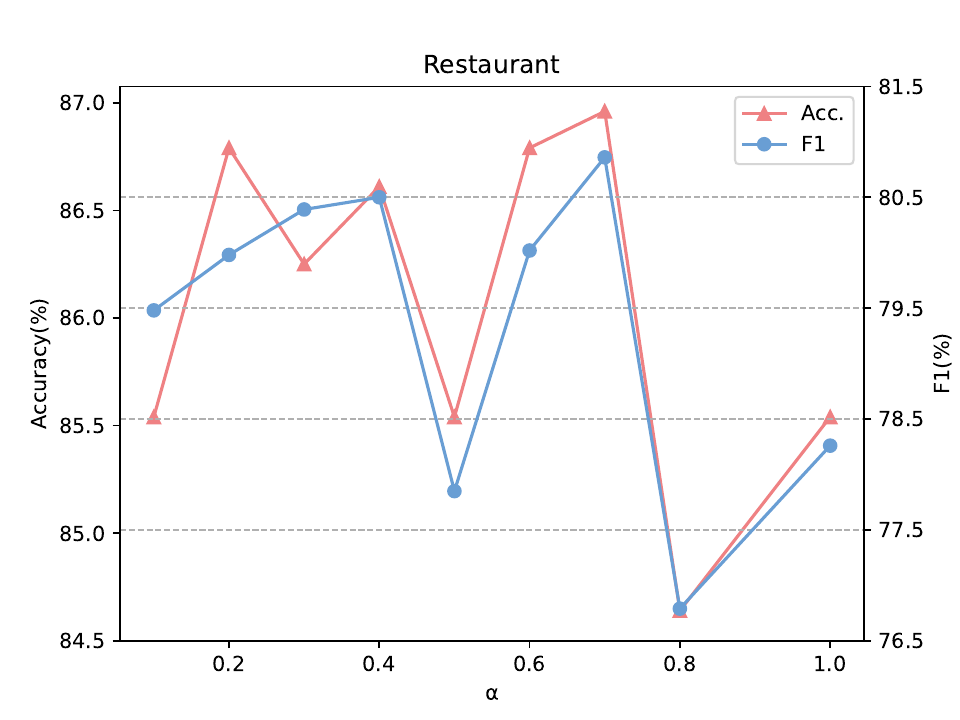}
}
\quad
\subfigure[\scriptsize{Res-CADA}]{
\includegraphics[width=0.92in]{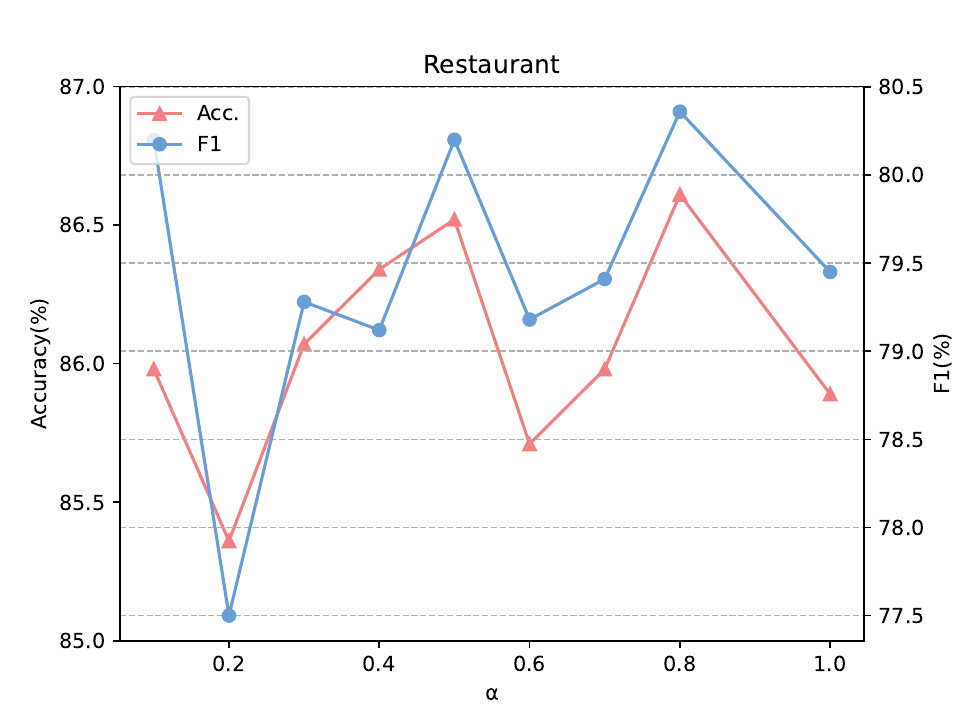}
}
\quad
\subfigure[\scriptsize{Res-CADA(veri)}]{
\includegraphics[width=0.92in]{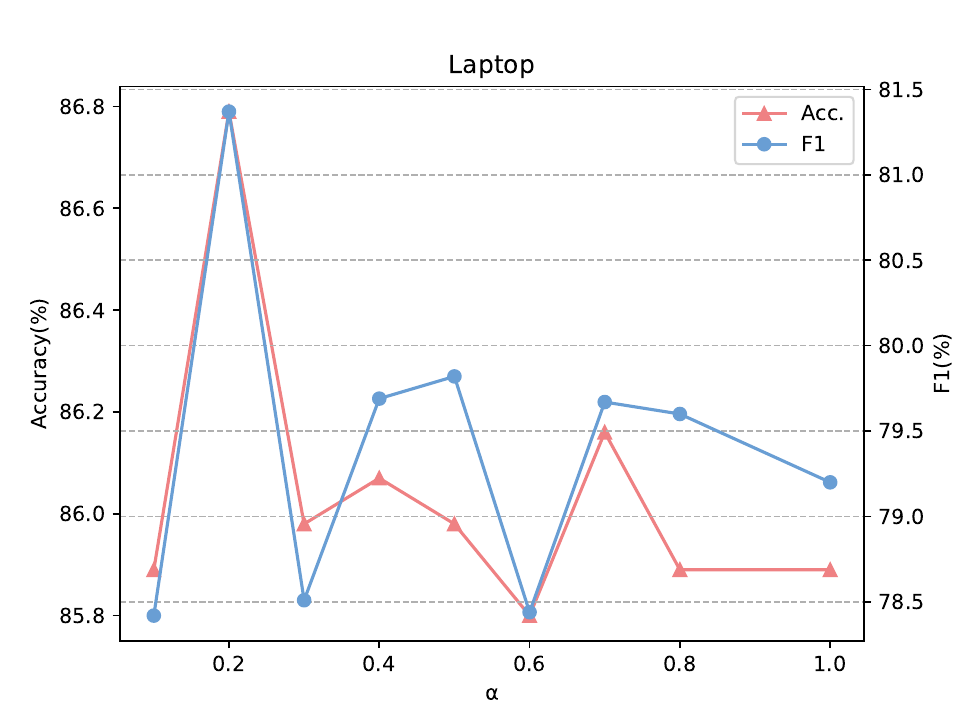}
}
\quad
\subfigure[\scriptsize{Lap-CDA}]{
\includegraphics[width=0.92in]{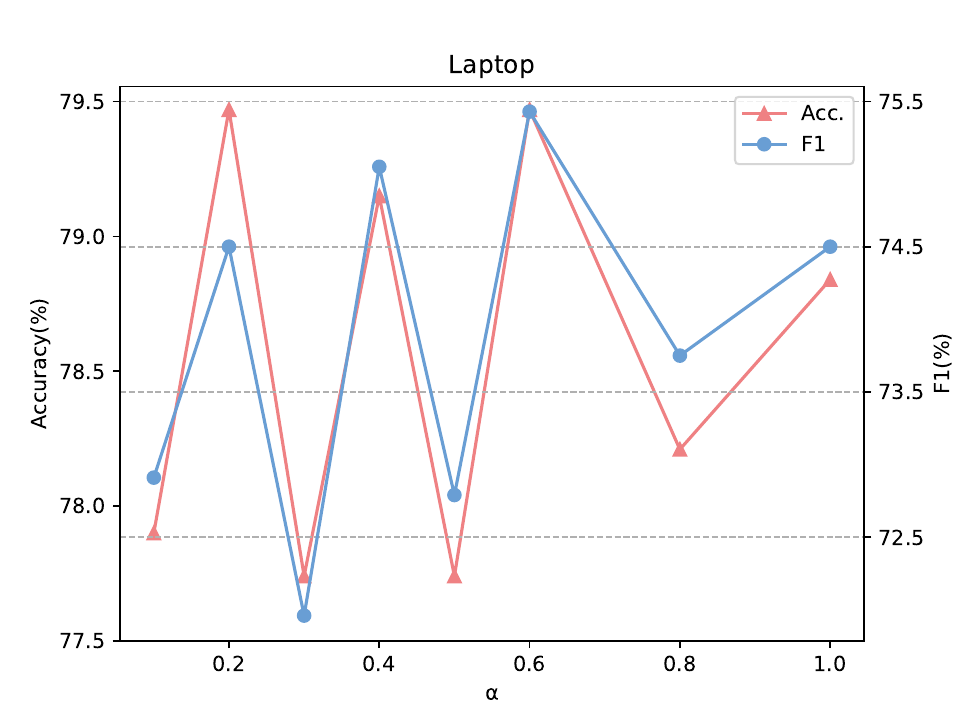}
}
\quad
\centering
\subfigure[\scriptsize{Lap-ADA}]{
\includegraphics[width=0.92in]{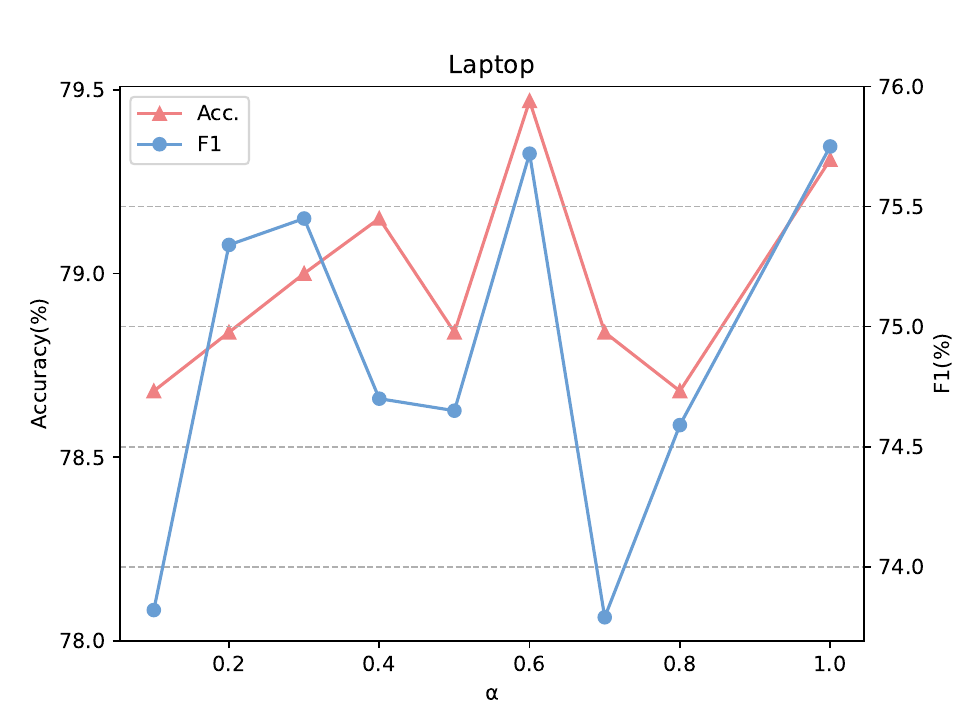}
}
\quad
\centering
\subfigure[\scriptsize{Lap-ADA (veri)}]{
\includegraphics[width=0.92in]{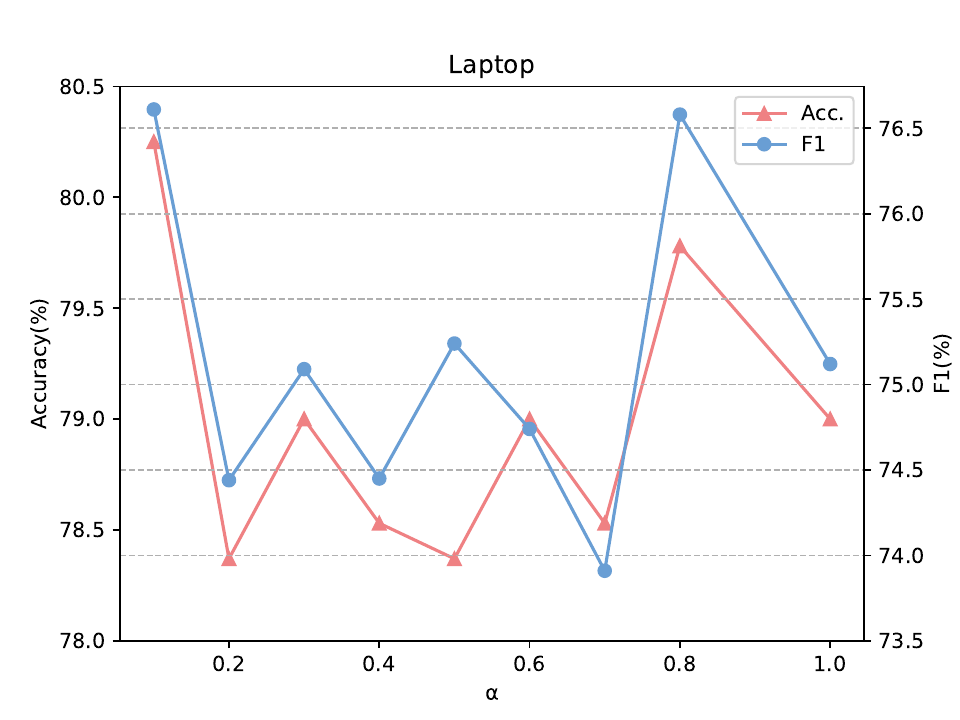}
}
\quad
\subfigure[\scriptsize{Lap-CADA}]{
\includegraphics[width=0.92in]{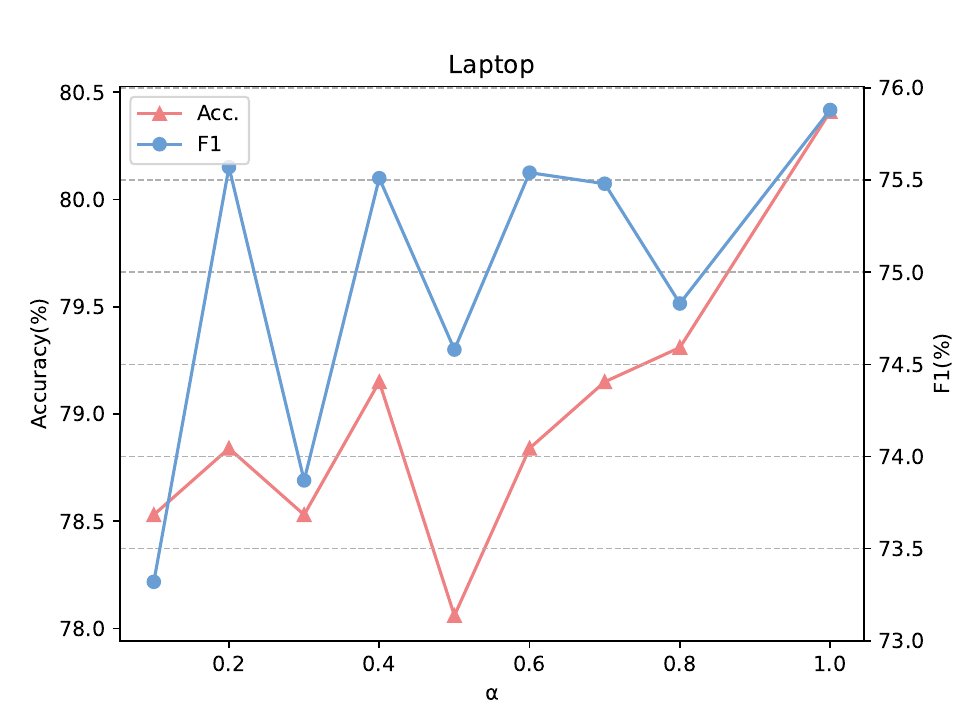}
}
\quad
\centering
\subfigure[\scriptsize{Lap-CADA(veri)}]{
\includegraphics[width=0.92in]{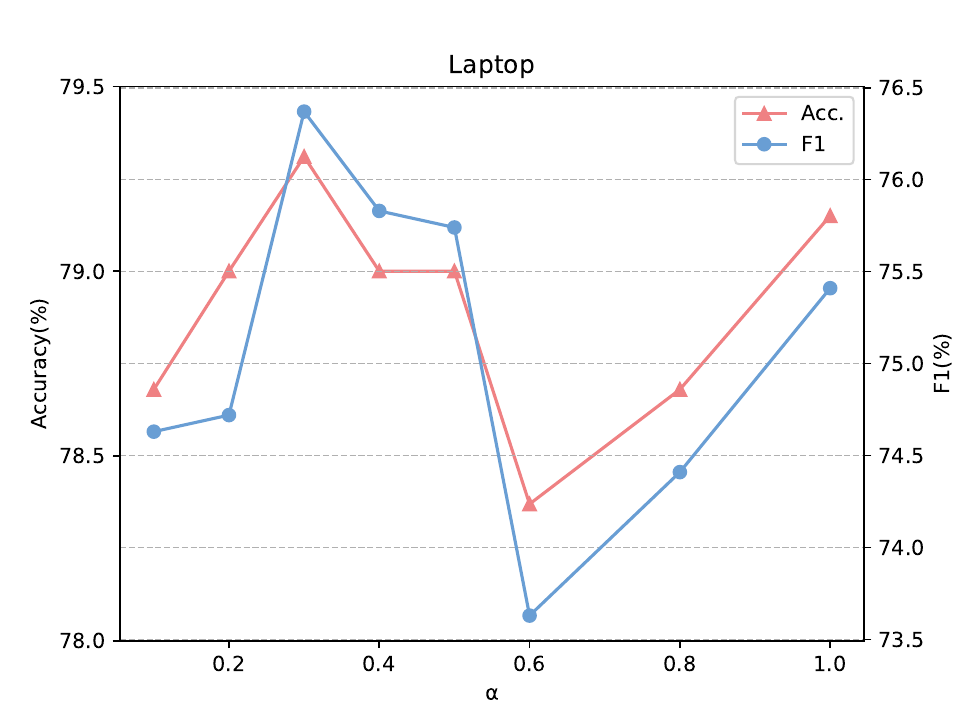}
}
\quad
\caption{Sensitivity analysis of the augmented cross-entropy loss hyper-parameter $\alpha$ in supervised aspect-based sentiment classification under different data augmentation strategies.}
\label{fig2}
\end{figure*}

\begin{figure*}
\centering
\subfigure[\scriptsize{Res-CDA}]{
\includegraphics[width=0.92in]{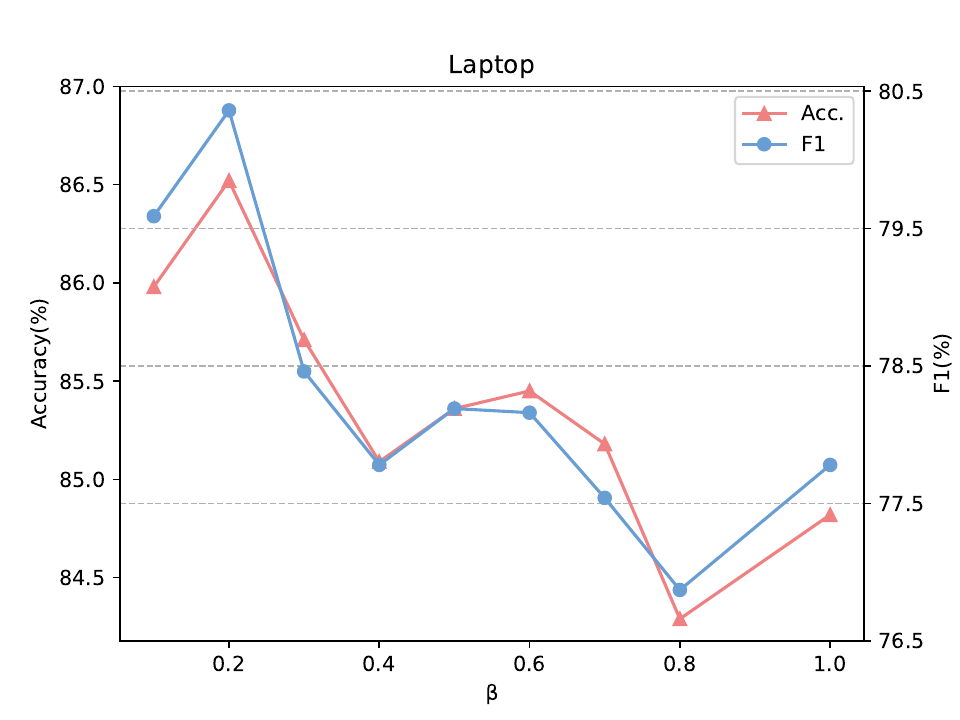}
}
\quad
\subfigure[\scriptsize{Res-ADA}]{
\includegraphics[width=0.92in]{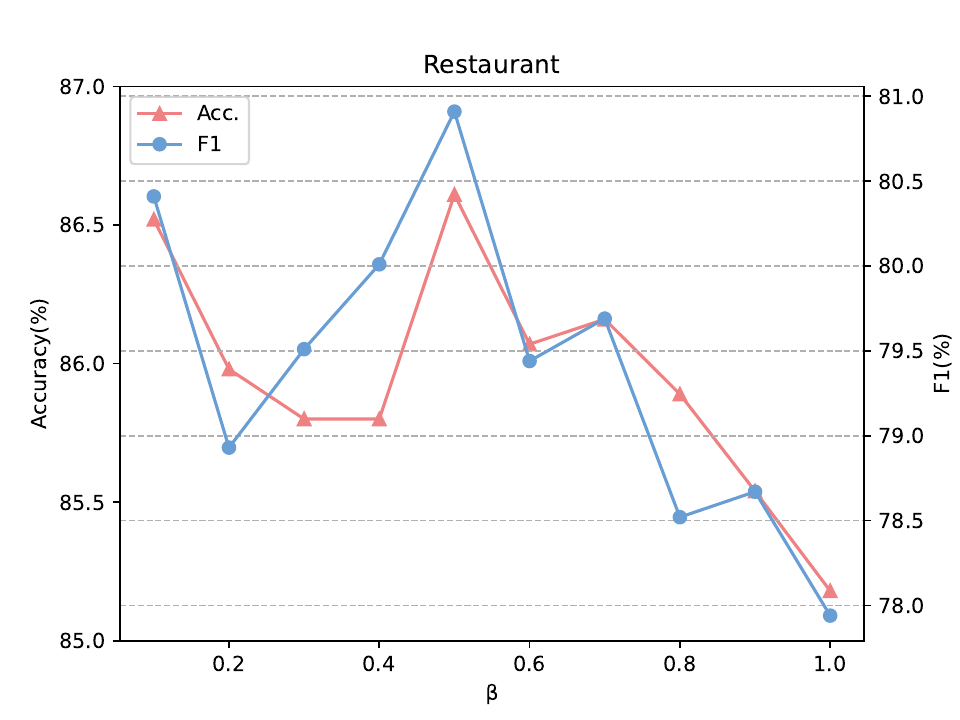}
}
\quad
\subfigure[\scriptsize{Res-ADA (veri)}]{
\includegraphics[width=0.92in]{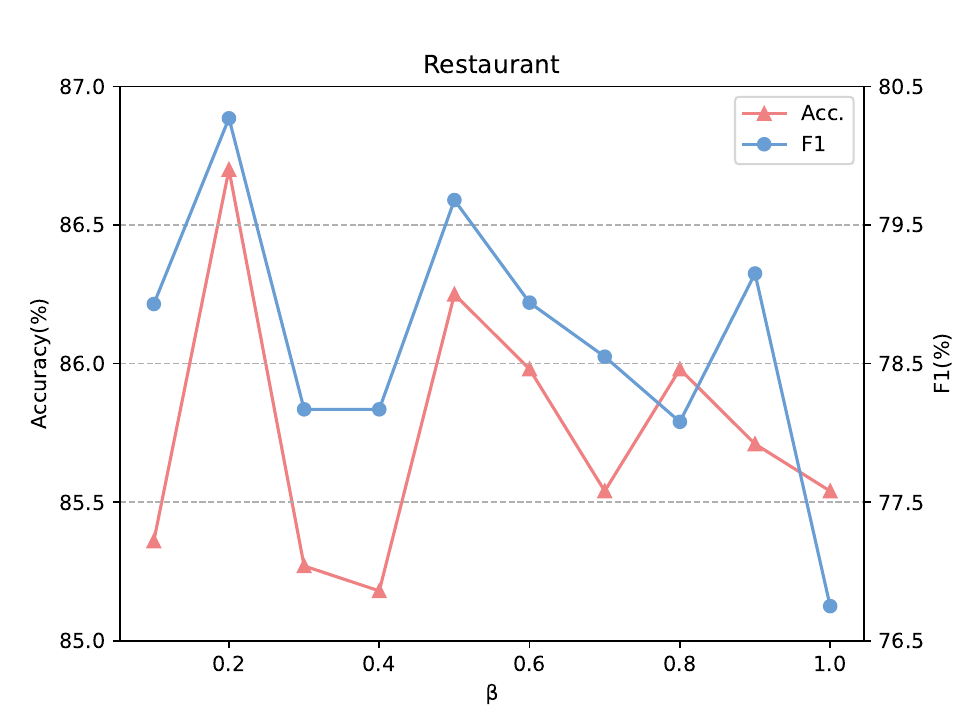}
}
\quad
\subfigure[\scriptsize{Res-CADA}]{
\includegraphics[width=0.92in]{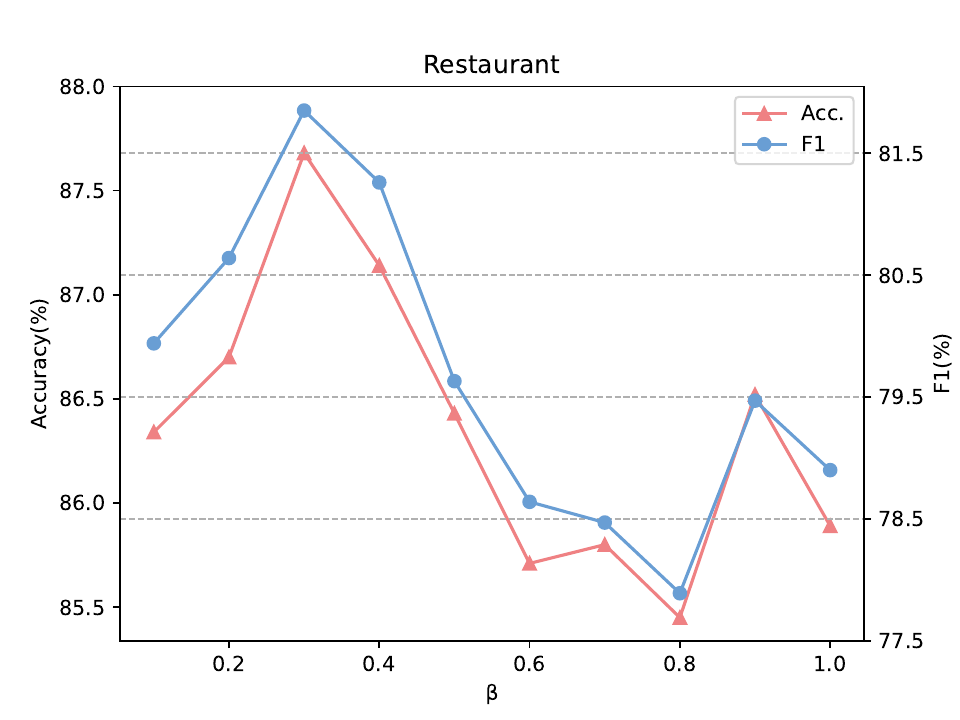}
}
\quad
\subfigure[\scriptsize{Res-CADA(veri)}]{
\includegraphics[width=0.92in]{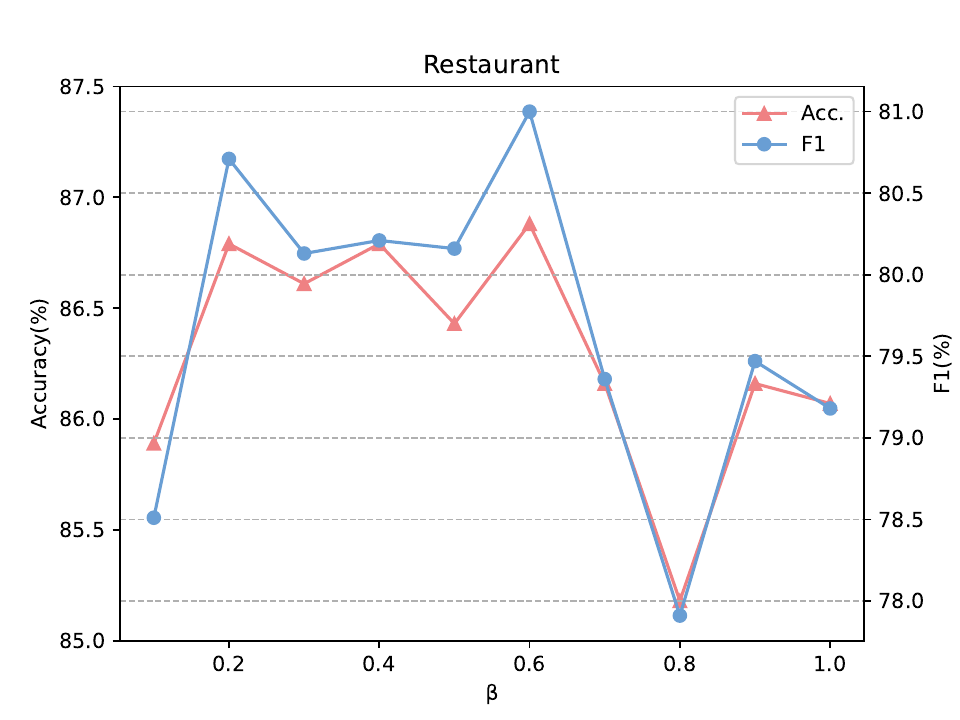}
}
\quad
\subfigure[\scriptsize{Lap-CDA}]{
\includegraphics[width=0.92in]{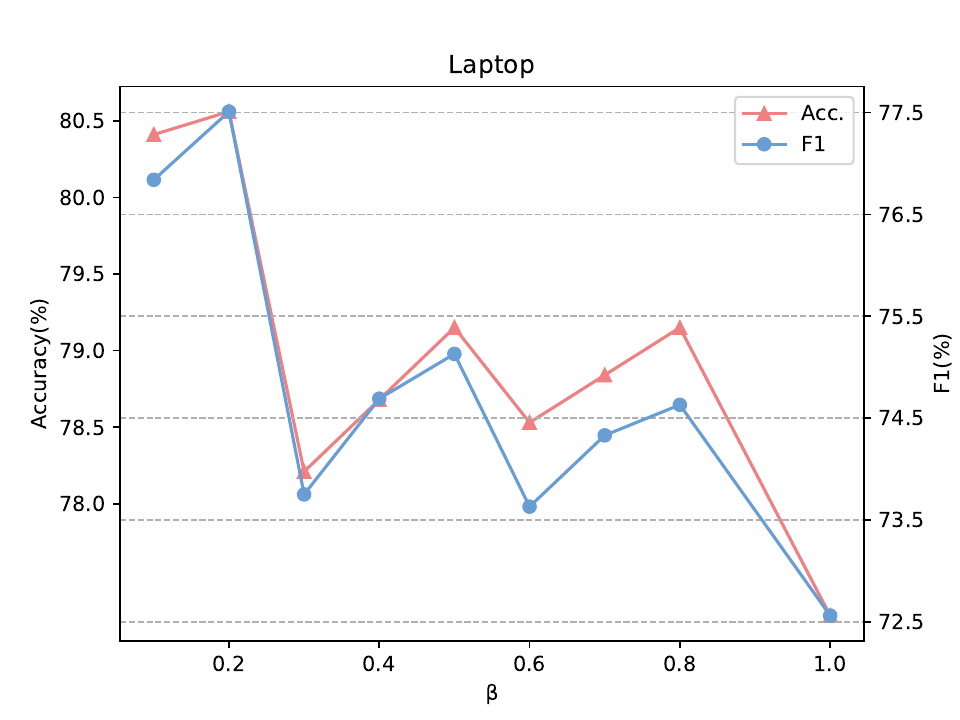}
}
\quad
\centering
\subfigure[\scriptsize{Lap-ADA}]{
\includegraphics[width=0.92in]{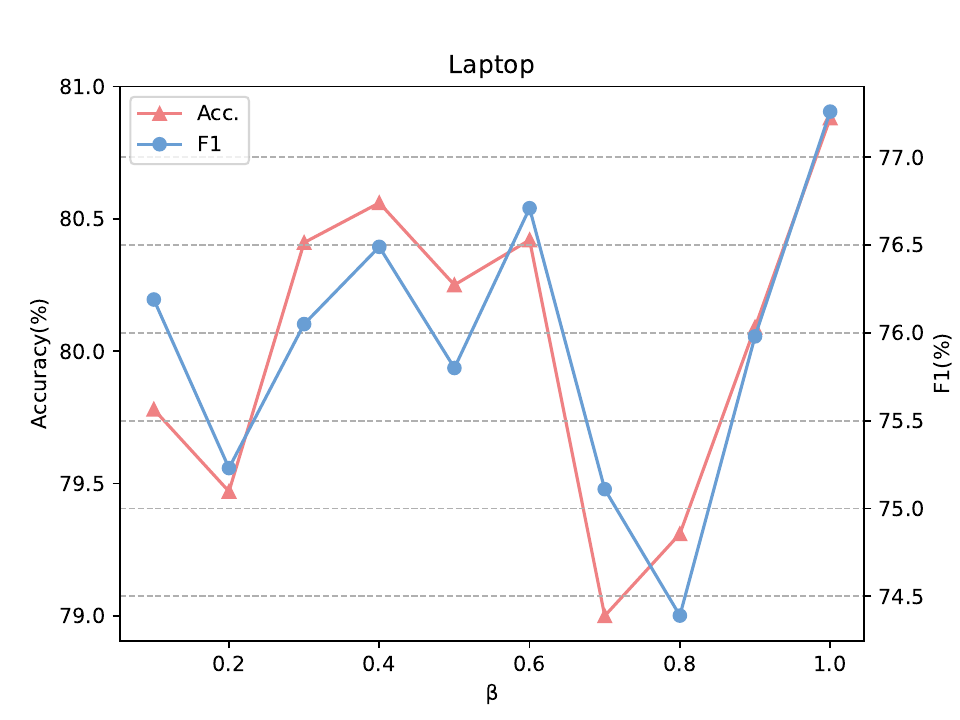}
}
\quad
\centering
\subfigure[\scriptsize{Lap-ADA (veri)}]{
\includegraphics[width=0.92in]{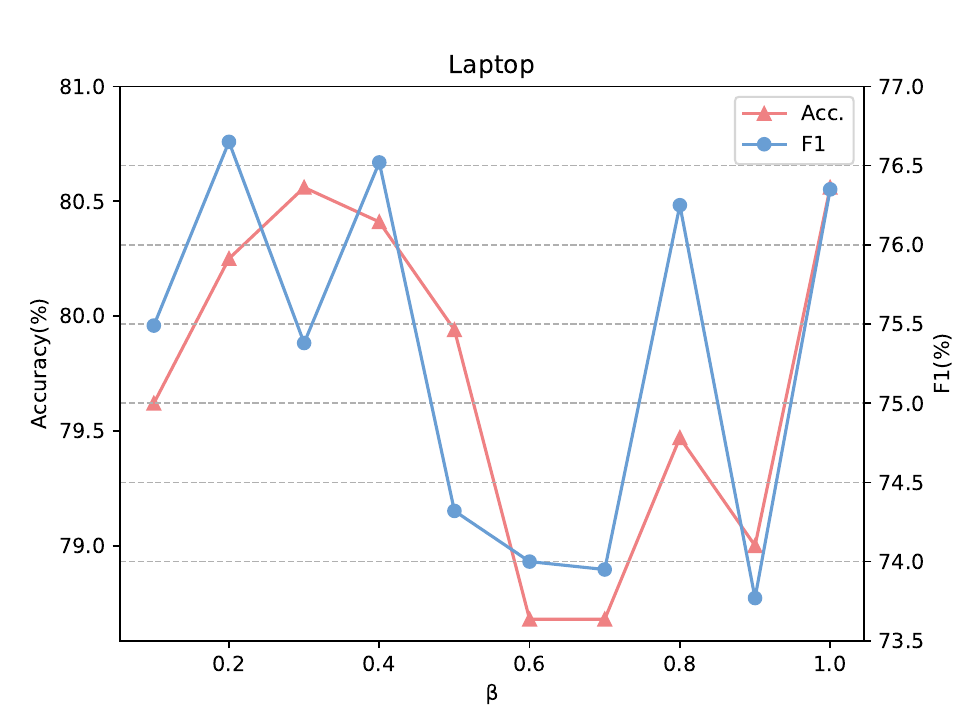}
}
\quad
\subfigure[\scriptsize{Lap-CADA}]{
\includegraphics[width=0.92in]{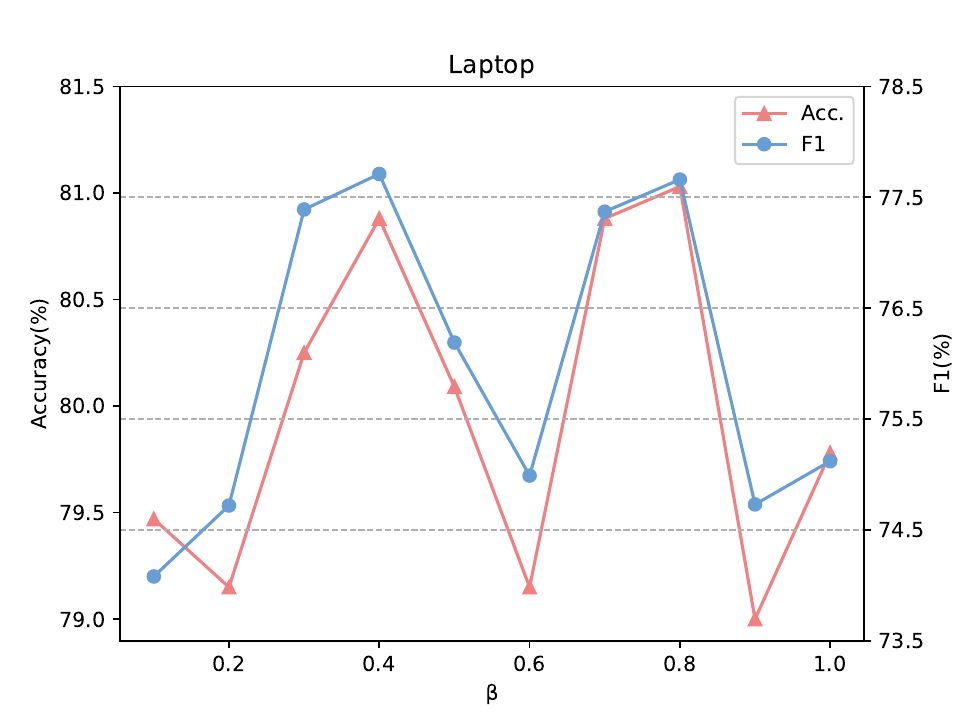}
}
\quad
\centering
\subfigure[\scriptsize{Lap-CADA(veri)}]{
\includegraphics[width=0.92in]{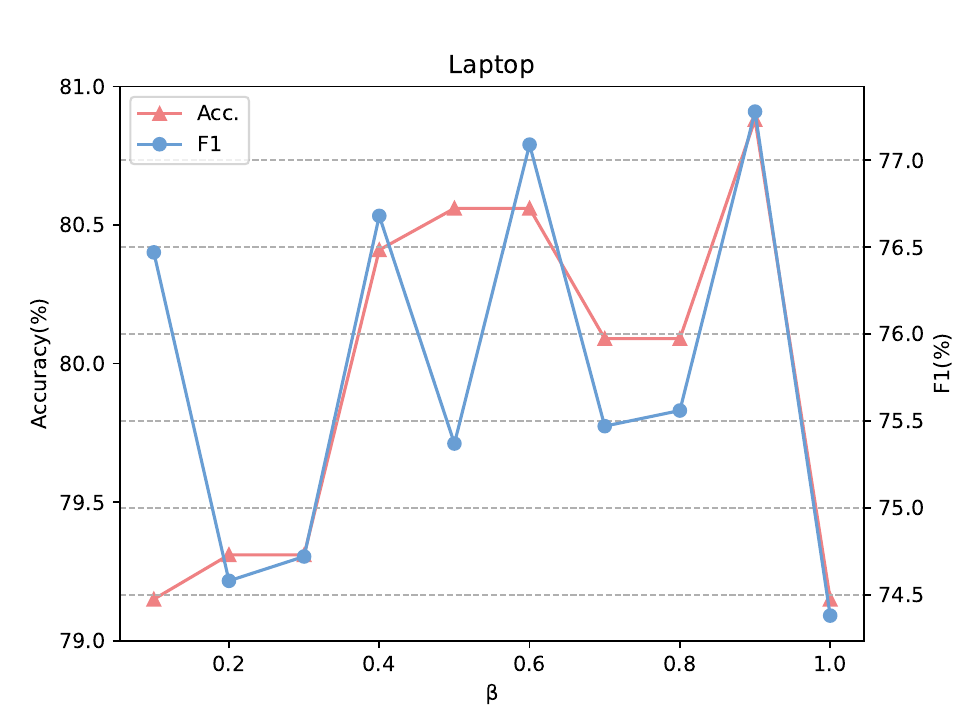}
}
\quad
\caption{Sensitivity analysis of the contrastive learning hyper-parameter $\beta$ in total training objective with three data augmentation strategies and data verification.}
\label{fig3}
\end{figure*}

\section{Conclusion}
In this paper, we investigated three data augmentation methods based on ChatGPT, combined with contrastive learning, to enhance the performance of ABSA tasks. The experimental results demonstrate that all three data augmentation strategies yield notable improvements in ABSA performance, with CADA exhibiting the best performance. The proposed method, CADA, not only enriches the semantic diversity of context words, but also improves the diversity of aspect terms, resulting in significant enhancements in ABSA performance that surpass baseline models by a substantial margin. The experimental results provide strong evidence for the effectiveness of our proposed framework, underscoring ChatGPT as a powerful data augmentation strategy that generates high-quality data through carefully designed prompts.

\section{Acknowledgements}
Qin's work was supported by a grant from the Research Grants Council of the Hong Kong Special Administrative Region, China (R1015-23). Xu, Xie and Wang's work was supported by a grant from the Research Grants Council of the Hong Kong Special Administrative Region, China (UGC/FDS16/E17/23); and Xie's work was also supported by the Faculty Research Grant (DB24C5) of Lingnan University, Hong Kong.

\bibliographystyle{acsa}
\bibliography{acsa}

\begin{IEEEbiography}{Lingling Xu}{\,} is currently pursuing a Ph.D.~degree at the School of Science and Technology, Hong Kong Metropolitan University. Her research interests include aspect-based sentiment analysis, and contrastive sentence representation learning. Contact her at \href{mailto:s1289650@live.hkmu.edu.hk}{s1289650@live.hkmu.edu.hk}.
\end{IEEEbiography}

\begin{IEEEbiography}{Haoran Xie}{\,} is currently the Acting Associate Dean and Professor at the School of Data Sciences, Lingnan University, Hong Kong. His research interests include artificial intelligence, big data, and educational technology. He has 414 research publications in international journals and conferences. Haoran Xie is the corresponding author of this article. Contact him at \href{mailto:hrxie@ln.edu.hk}{hrxie@ln.edu.hk}.
\end{IEEEbiography}

\begin{IEEEbiography}{S. Joe Qin}{\,} is currently the Wai Kee Kau Chair Professor and President of Lingnan University, Hong Kong. His research interests include data science and analytics, machine learning, process monitoring, and model predictive control. He has over 470 publications in international journals and conferences. Contact him at \href{mailto:joeqin@ln.edu.hk}{joeqin@ln.edu.hk}.
\end{IEEEbiography}

\begin{IEEEbiography}{Fu Lee Wang}{\,} is currently the Dean of the School of Science and Technology, Hong Kong Metropolitan University, Hong Kong. His research interests include educational technology, information retrieval, computer graphics, and bioinformatics. He has over 300 publications in international journals and conferences. Contact him at \href{mailto:pwang@hkmu.edu.hk}{pwang@hkmu.edu.hk}.
\end{IEEEbiography}

\begin{IEEEbiography}{Xiaohui Tao}{\,} is currently the School Head, Acting Dean, and Professor of School of Mathematics, Physics, and Computing, University of Southern Queensland, Australia. His research interests include artificial intelligence, knowledge engineering, and health informatics. He has more than 150 publications in international journals and conferences. Contact him at \href{mailto:xtao@usq.edu.au}{xtao@usq.edu.au}.
\end{IEEEbiography}

\end{document}